\documentclass{article}

\usepackage{arxiv}

\usepackage[utf8]{inputenc} 
\usepackage[T1]{fontenc}    
\usepackage{hyperref}       
\usepackage{url}            
\usepackage{booktabs}       
\usepackage{amsfonts, textcomp}       
\usepackage{nicefrac}       
\usepackage{microtype}      
\usepackage{lipsum}
\usepackage[colorinlistoftodos]{todonotes}
\usepackage{graphicx}
\graphicspath{ {./Figures/} }

\title{Automated decontamination of workspaces using UVC coupled with occupancy detection}

\author{
\begin{tabular}{cc}
{\bf Asit K Mishra}  & {\bf Federico Tartarini} 
\tabularnewline 
 \shortstack{Berkeley Education Alliance for \\ Research in Singapore} & \shortstack{Berkeley Education Alliance for \\ Research in Singapore} \tabularnewline
 \texttt{writeto.asit@gmail.com}& \texttt{federico.tartarini@bears-berkeley.sg} \tabularnewline
  & \tabularnewline
 {\bf Zuraimi Sultan} & {\bf Stefano Schiavon} \tabularnewline
  \shortstack{Berkeley Education Alliance for \\ Research in Singapore} & University of California, Berkeley \tabularnewline
\texttt{zuraimi.sultan@bears-berkeley.sg} &  \texttt{schiavon@berkeley.edu} \tabularnewline
\end{tabular}
}

\begin{document}
\maketitle

\begin{abstract}
Periodic disinfection of workspaces can reduce SARS-CoV-2 transmission.
In many buildings periodic disinfection is performed manually; this has several disadvantages: it is expensive, limited in the number of times it can be done over a day, and poses an increased risk to the workers performing the task. 
To solve these problems, we developed an automated decontamination system that uses ultraviolet C (UVC) radiation for disinfection, coupled with occupancy detection for its safe operation. 
UVC irradiation is a well-established technology for the deactivation of a wide range of pathogens. 
Our proposed system can deactivate pathogens both on surfaces and in the air. 
The coupling with occupancy detection ensures that occupants are never directly exposed to UVC lights and their potential harmful effects. 
To help the wider community, we have shared our complete work as an open-source repository, to be used under GPL v3.
\end{abstract}


\section{The Covid-19 Pandemic: Space disinfection with UVC light}

The Covid-19 pandemic has changed the way we live in, design and operate buildings. Even when this pandemic will be over, there could always be similar other pathogens, with pandemic potential. We are also continuously at risk from antimicrobial-resistant microbes, which World Health Organization (WHO) considers one of the top ten public health threats~\cite{who_antimicrobial_2020}. To ensure that future buildings are more resilient to such threats, we need to consider non-pharmaceutical interventions (NPIs) that can make buildings safer for occupants. Ultraviolet radiation C, 200-280~nm wavelength, also known as UVC, is an NPI that can be used to decontaminate indoor surfaces and air of multiple pathogens.  

UVC damages the genetic material of cells, thus inactivating unicellular pathogens. Cellular nucleic acids, DNA for most cells and RNA in case of certain viruses, strongly absorb UVC radiation, particularly in the wavelength range of 250 to 270~nm, making cellular replication unviable~\cite{dai_ultraviolet_2012}.
Because biological tissues are prone to absorbing UVC, its use poses a threat mainly to unicellular organisms and the first cellular layer of exposed skin and eyes in humans~\cite{ashrae_ultraviolet_2015}. 

UVC lights are already widely used for decontamination in specialized circumstances like hospitals and research laboratories~\cite{doll_touchless_2015, otter_role_2013}. But the pandemic has brought the focus back on potential use of UVC in other building types~\cite{NUNAYON2020122715}. 
Some recognized advantages of decontamination using UVC are that it has short cycle times (minutes), it is easy to operate, it can be easily adapted to fully automated operation, and spaces do not need to be sealed and occupants can immediately return to their workplace after decontamination. 
When used as per advised protocols, it does not produce any undesirable by-products --- like other chemical based disinfectants --- in the space~\cite{otter_role_2013}.
UVC lights can, however, degrade some materials (particularly, polymeric materials)~\cite{ashrae_ultraviolet_2015}

SARS-CoV-2, the virus that causes Covid-19, can be transmitted by asymptomatic and presymptomatic individuals~\cite{moghadas_implications_2020}. Such individuals can unknowingly infect other occupants, when in close contact indoors~\cite{bulfone_outdoor_2020}. Hence, there is a need for focus on indoor spaces because of higher risks and possible, close range interaction with people who are not presenting symptoms yet. 
To reduce the risk that building occupants get infected indoors, current practices for workspaces advise disinfection of touch surfaces using chemical disinfectants. This approach, however, is expensive, can be performed only limited number of times during a workday, and exposes the cleaning crew to undue risk. To overcome the aforementioned limitations, manual surface cleaning can be replaced or conducted in parallel with the use of automated UVC cleaning. What hinders the use of UVC to decontaminate indoor spaces, is the potential damage that prolonged exposure of UVC irradiation, over several minutes, can cause to skin and eye~\cite{ashrae_ultraviolet_2015}. This restricts UVC disinfection to be carried out only in absence of occupants. However, in the current situation, workspaces that want to operate close to their normal capacity and take advantage of decontamination using UVC, cannot relegate such cleaning to after office hours. They need decontamination to be carried out multiple times through the work-day.

To address this problem, we developed and tested an automated disinfection system that uses UVC lights coupled with an advanced occupancy detection system. The latter comprises of multiple layers of safety. Our system is automatically activated, multiple times throughout the day, whenever the space is not occupied. It can disinfect both surfaces and deactivate pathogens suspended in the air. 
Hence, it reduces the number of times indoor spaces need to be manually cleaned, minimizing cleaners' exposure to active pathogens.

\section{System Description and Case Study Design}
Our system includes three UVC disinfection tiers: upper room, ceiling lights, and a desk light. 
The upper room UV system is installed at a height of 2.4~m. 
It can be turned on at all time, and safely operated while the space is fully occupied~\cite{miller_upper_2015}, without risk to people~\cite{first_monitoring_2005, niosh_environmental_2009}. 
The latter two, on the other hand, are facing downwards, consequently the occupants can be exposed to the UVC when they are turned on.
Our objective was that both the desk light and the ceiling lights should automatically turn off when anyone comes in their close proximity. 
Consequently, for safety, multiple layers of occupancy detection were used to ensure we do not encounter false negatives, i.e., the system perceives that the space is unoccupied while in reality, a person is present. 
To achieve this, we relied on the following technologies:

\begin{enumerate}
	\item A combination of passive infrared (PIR) and ultrasonic (US) occupancy sensors, functioning together to turn off ceiling and desk light when occupancy is detected.
	\item Bluetooth Low Energy (BLE) beacons are used for sensing occupants approaching the space. 
	When any occupant is perceived to be closing in, based on the signal strength, the ceiling lights are automatically turned off. 
	This ensures that the UV lights are turned off even before an occupant enters the space.
	\item A manual switch can be used to turn off all the lights if all the other safety layers fail.
\end{enumerate}

We used two PIR sensors, mounted at the two corners of the room close to the desk lamp. 
Each sensor has a 120\textdegree field of view both in the horizontal and vertical direction and together, they covered the entire floor area. 
The US sensor was pointing to the location where the occupant  was seated at the desk with the desk lamp. 
Its detecting distance was adjusted to 2~m.

We tested our system in the state-of-the-art testbed owned and operated by the Berkeley Education Alliance for Research in Singapore (BEARS). 
The room has a floor area of 25~m\textsuperscript{2} (width 4.3~m and length 5.6~m) and ceiling height of 2.6~m. 
There were two desks placed at each end of the room, 4.4~m apart. 

In this manuscript we report only the results of the tests that we conducted related to safety and functional effectiveness of the system. 
Our primary aim, from a safety point of view, was that the system was able to turn off automatically the UVC lights before occupants may have exposed to direct UVC radiation. 
One of our objectives was also to determine for how long the lights needed to be operational, after the last occupant left the space, to ensure that the minimal dose of electromagnetic radiation required for deactivating viruses similar to SARS-CoV-2 was delivered. 
From the available literature, this dosage is estimated to be between 20-27~J/m\textsuperscript{2}, for 90\% deactivation, and is referred to as D90~\cite{wladyslaw_j_kowalski_2020_2020, storm_rapid_2020, beggs_upper-room_2020}. 
Since UVC lights can degrade some materials they were controlled to be operational only as much as needed. 
We initially set the upper room UVC to be operational for 5 minutes every hour, while ceiling lights for 10 minutes and desk light for 5 minutes once turned on, unless their operation was interrupted by an occupant entering the space and forcing the lights to turn off. 
These operational times can be subsequently changed by the system operator. 
We chose these values based on lamp specifications provided by manufacturers and the time estimated to reach the D90 dosage.

The algorithm is programmed to ensure that once all occupants leave for the day, all UVC lights are kept on for one such cycle. 
Lights are again turned on at midnight for one cycle and then kept off until occupancy is again detected, to avoid unnecessary operation during weekends and holidays.
As previously mentioned, the upper room UVC does not impact occupants and could be kept on safely, even when occupants are present. 
However, we decided to keep it on only for 10 minutes. 
The focus of our work on safe operation of the lamps focused on occupied spaces (ceiling and desk light) and we did not want to rehash a known, safe technology like the upper room UVC light.

The deployment used the following UVC luminaires: 
\begin{enumerate}
    \item One Desk UVC lamp: 24~W, Osram - HNS L 24 W 2G11, 253.7~nm.
    \item Two ceiling UVC lamps: 36~W, Signify - UVC Batten, 253.7~nm.
    \item One upper room UVC lamp: 25~W, Signify - APF-WM, 253.7~nm.
\end{enumerate}

\begin{figure}[!ht]
	\centering
	\includegraphics[width=\textwidth]{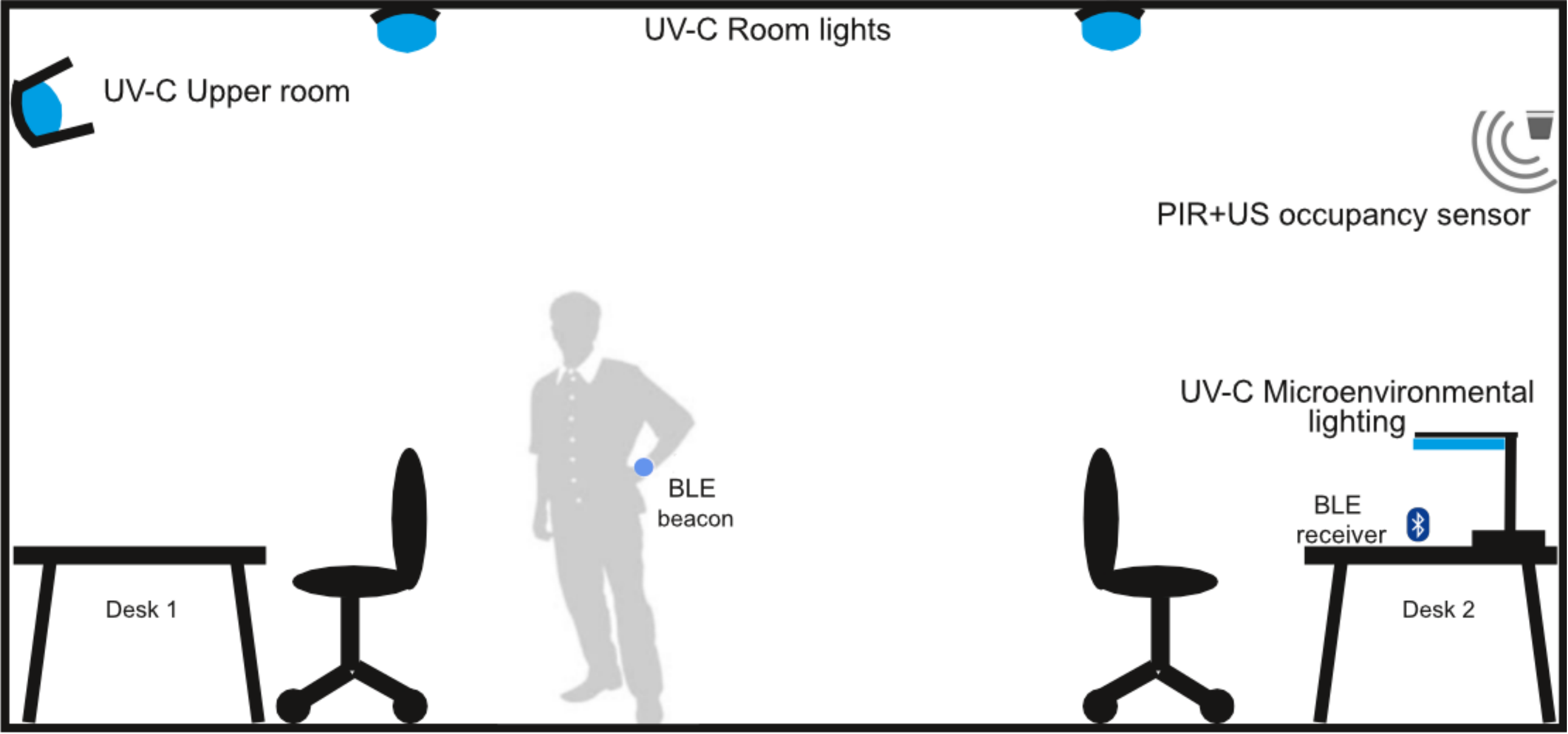}
	\caption{A schematic of the automated UVC disinfection system's techniques for the safe set-up and operation of the UVC lamps. 
	Schematic not intended as an exact replica.}
	\label{Fig1}
\end{figure}

\begin{figure}[!ht]
	\centering
	\includegraphics[width=\textwidth]{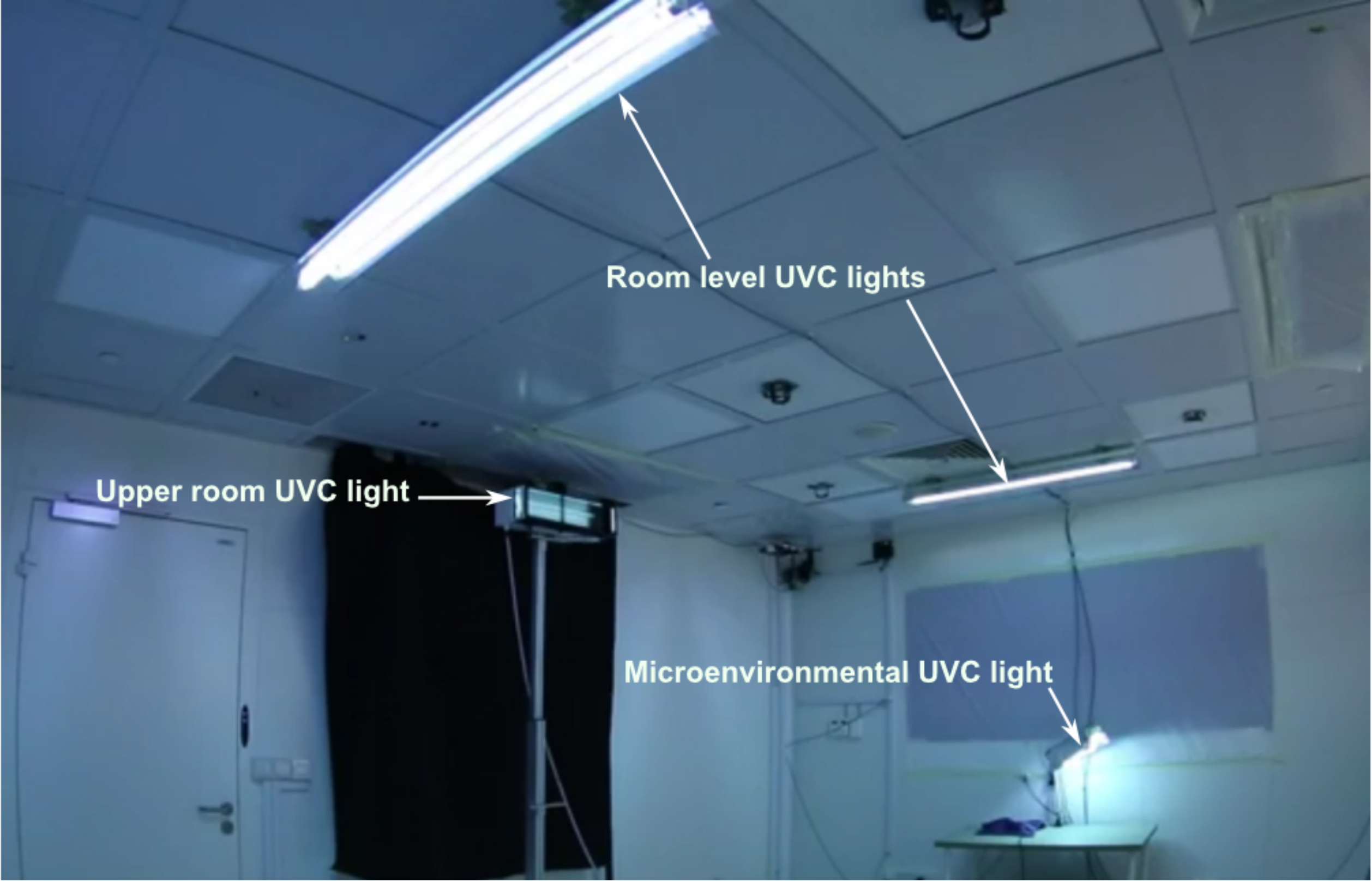}
	\caption{UVC lamps installed in the testbed room during the experiment.}
	\label{Fig2}
\end{figure}

A schematic of the room set-up is provided in Figure~\ref{Fig1} and an image of the three types of lamps, in operation, has been presented in Figure~\ref{Fig2}.

In our experiment, we installed only one desk lamp in one workstation desk (i.e., Desk~2), as shown in Figure~\ref{Fig1}. 
This was to test the situation that a desk lamp can be turned on even with someone working from the other workstation desk (i.e., Desk~1). 
Since there are both desk and ceiling lights, we performed the following tests.
Each lasted for a 2-hour period.

\textbf{Test BLE beacon detection}. 
Occupant (experimenter) seated at Desk~1, while carrying a BLE beacon in his pocket. 
With the BLE being detected, the ceiling lights should stay off. 
Being the occupant seated sufficiently far from the other desk and not being exposed to direct UVC light from Desk~2 lamp, this could turn on.

\textbf{Test occupant movement and BLE beacon detection}. Occupant seated at Desk 2, while carrying a BLE beacon in his pocket. 
Both the ceiling and the desk lamps must stay off at all times.

\textbf{Test of UVC operation, without occupants}. 
Neither occupants nor beacons in the room. 
After the last occupant leaves the room, the lights should turn on for one cycle and then stay off.

\textbf{Test occupant movement detection}. 
Occupant transitioning across and around the room. 
After occupant leaves the room, lights turn on for one cycle. 
Subsequently, when the occupant comes back to the space, if lights are on they immediately turn off and later, after the occupant leaves, turn on again for one cycle.

During the safety tests, UVC irradiance was measured in two locations: on Desk~2 and in the centre of the room, at a height of 0.7~m (desk level height) using Lutron UVC Light Meter (Model UVC-254A, accuracy: $\pm$~2\% full scale) with a logging frequency of six times every minute. 
After conducting the safety testing, average UVC irradiance, with all lights on, was measured on the floor of the room, over a 6$\times$8 = 48 points grid, using the same sensor. 
The floor was chosen for these measurements since it is farthest from the ceiling mounted lamps and irradiance intensity decreases as the inverse of distance squared.
These results were used to determine for how long each light cycle should last to reach the D90 dosage.

We uploaded the source code, the schematic of the controller, and the documentation needed to install the source on a RaspberryPi,  in a public GitHub repository: \url{https://github.com/CenterForTheBuiltEnvironment/rpi-uv-controller}. 
We released the source code under a GNU General Public License v3.

\section{Results From the Case Study and Future Perspectives}
We present in this section the results of the safety test and the results for the UVC irradiance measured on the floor. 
From safety perspective, while we can accept false positives --- the system detecting movement when no one is inside the room --- we cannot accept false negatives due to the negative health consequences. 
From UVC irradiance perspective, we expected that at floor level, at least D90 dosage should be achieved within the 10~minute operational cycles planned.

\begin{figure}[!ht]
	\centering
	\includegraphics[width=\textwidth]{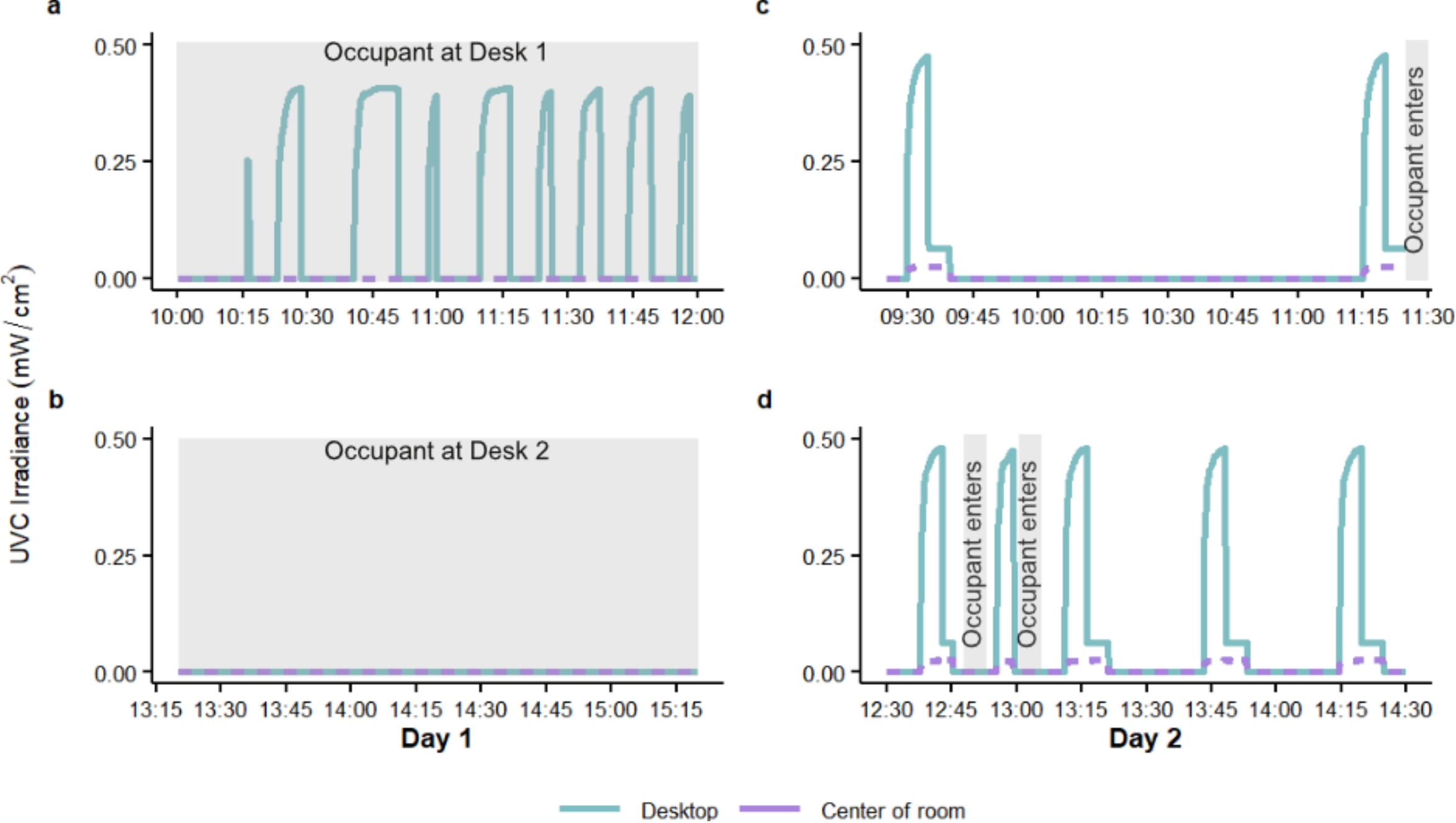}
	\caption{Safety testing of UVC lamps operation as measured from the UVC irradiance recorded at the center of the room and  Desktop~2. 
	Occupant presence is indicated using a grey background. 
	a) Test of ceiling light safety and desk operation 
	b) Test of safety --- both levels --- with occupant present c) Test of UVC operation, without occupants 
	d) Test of safety with occupant movement.}
	\label{Fig3}
\end{figure}

Figure~\ref{Fig3} provides the UVC irradiance recorded at the center of the room and on top of Desk~2, both measured at a height of 0.7~m. 
Figure~2~a) shows that desk light can be turned on with occupant working at Desk~1.
The UVC irradiance at room centre remained zero, implying no risk to someone at Desk~1 from the desk lamp installed on the other desk. 
Both ceiling lights stayed off during this period. 
Movements of a person at Desk~1, like, rolling the chair back or getting up from their chair, was detected by the PIR sensor.
Since a person seated at Desk~1 is far enough from Desk~2, subsequent to a time gap after occupancy is detected, the desk lamp could be triggered on. 
This behaviour is expected to ensure localised decontamination following people leaving their desk for some time. 
Figure~\ref{Fig3}~a) shows that the desk lamp was triggered on, following a gap after occupant presence was detected and off when occupant presence was again detected.
Because of the occupant at Desk~1 being intermittently detected by the PIR sensor, the lamp frequently does not complete its full cycle run of five minutes.
This allowed us to verify that when someone steps away from their desk, for example to talk with colleagues, the desk UVC lamp is turned on, without turning ceiling UVC lights on. 
And once occupant comes back close to the desk (Desk~2 in this instance), the desk UVC light is turned off.

Figure~\ref{Fig3}~b), with occupant at Desk 2, depicts that neither the ceiling nor desk lamps turned on, as expected.

Figure~\ref{Fig3}~c) shows the situation where no occupants were present in the room. 
One disinfection cycle was triggered after the occupant left the room, as expected. 
But, later during this period, there was a false positive, which triggered one more disinfection cycle. 
This cycle ended abruptly as the occupant approached the room, while carrying the BLE beacon. 
This shows that the BLE worked as per expectation. 

The data collected with the occupant moving across and around the room are shown in Figure~\ref{Fig3}~d). 
The first two operations were triggered, post detection of occupant presence, as desired. 
And they were also terminated when occupant presence was detected again.
Between 13:45 and 14:30 of Day 2, we do see two false positives triggering two more cycles of UVC operation. 
As part of safety tests, we verified that the system did not provide any false negatives (i.e., UVC lights off when an occupant was in the room), at either the room level or the desk level. It may also be noted from Figure~\ref{Fig3}~c) and d) that since the ceiling lamps had double the cycle time of the desk lamp, irradiance at desk level changes after the desk lamp is off and only ceiling lamps are on.

Measuring the irradiance on the floor of the chamber, over a grid of 48 points, allowed us to determine the time required to achieve the D90 dosage over the entire floor. 
The results are presented in Figure~\ref{Fig3}.
Being all the measured values lower than 300~s we deduced that operation of the lights for about five minutes lead to D90 dosage. 
Operation of 10 minutes leads to two cycles of D90 dosage, which is equivalent to 99\% disinfection or D99 ($0.99=0.90+0.90\times(1-0.90)$). 
The initially chosen cycle time for ceiling lights, was hence able to provide D99 dosage on almost the entirety of the room surfaces.
We originally predicted that with our estimated operational times we could reach a D90 dosage but it turned out, we can achieve a D99 with the given settings. 
Thus, a disinfection cycle can be successfully carried out even when occupants leave their desk for a short discussions or break. 
Considering that occupants in workspaces are likely to take several of such breaks over a work-day, multiple decontamination cycles can be performed. 
Depending on room size, orientation, surfaces and furniture where our system will be installed, the cycle times can be adjusted accordingly, by simply editing the relevant variables in the source code.

\begin{figure}[!ht]
	\centering
	\includegraphics[width=0.45\textwidth]{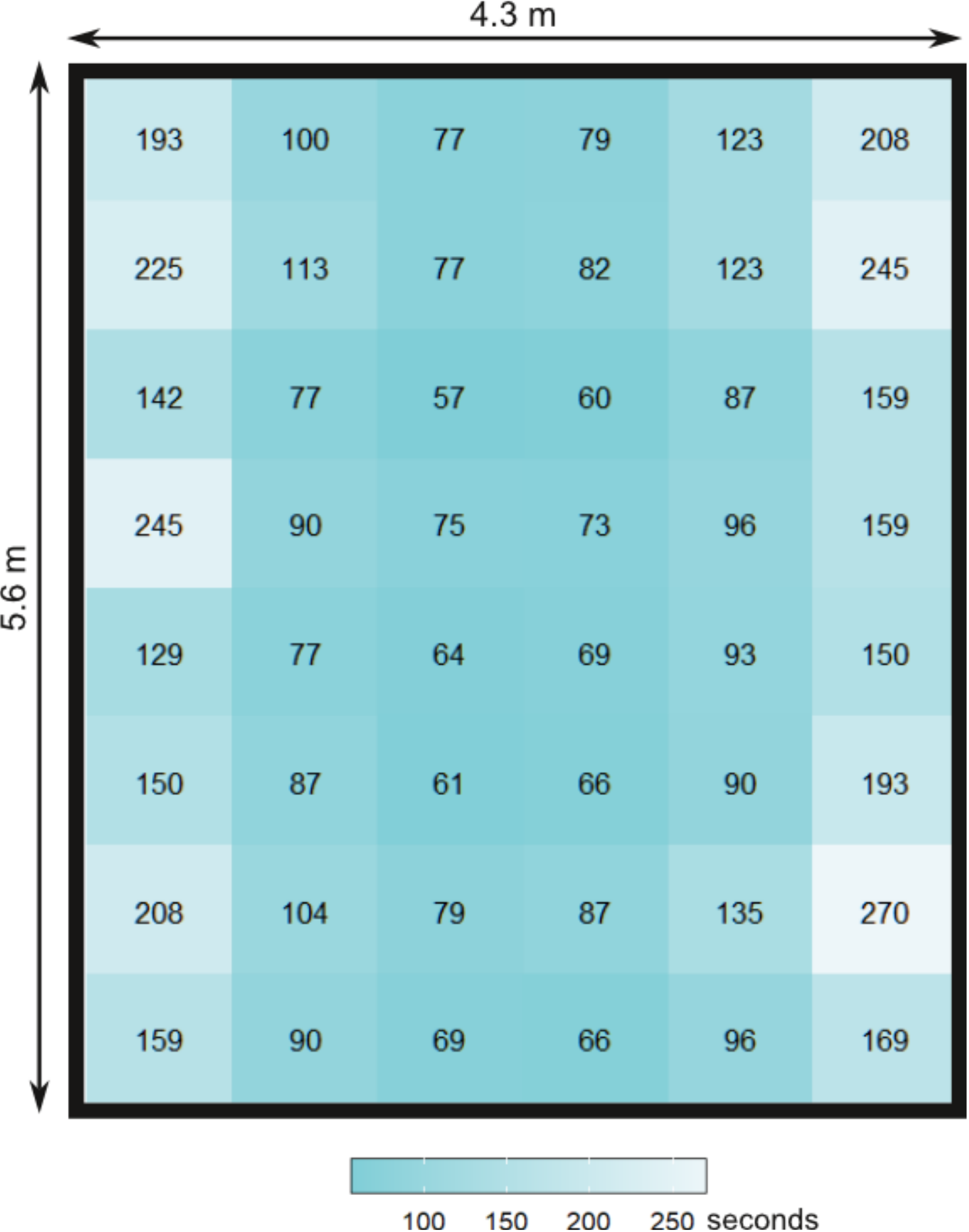}
	\caption{Time needed, in seconds, to achieve D90 dosage for SARS-CoV-2 on 48 grid points on the floor of the testbed room. Time calculated from measured UVC irradiance and assumed D90 of 27~J/m\textsuperscript{2}}
	\label{Fig4}
\end{figure}

As previously described our system relies on three separate layers of occupancy detection.
The PIR and ultrasonic occupancy sensors do not depend on active occupant cooperation. 
They can trigger both ceiling and desk UVC lights off upon detecting occupant presence in the space. 
But, unlike the beacon, they need an occupant to be inside the space. 
They cannot trigger ceiling lights off before an occupant enters the space. 
BLE beacons were detected by our micro-controller before the occupant entered the space, and the ceiling UVC luminaires were turned off while the occupant approached the space.   

We feel that for best results, occupants who work in a place where our system is installed must carry a BLE beacon for an extra layer of safety, at all times. 
While this requires an active cooperation from them and adds an extra layer of complexity, it ensures that people are never exposed to direct UVC radiation.
During out tests, we showed that carrying a BLE beacon would minimize any chance of exposure of occupants to the UVC radiation.

This pandemic has seen the use of contact tracing cell phone apps, using BLE technology, successfully, by countries like Singapore and Australia to trace and isolate possible contacts of infected individuals~\cite{abbas_covid-19_2020}, thus controlling spread. 
In balance, the benefits of using a BLE beacon look like they would outweigh the extra complexity and occupant cooperation expected. And with BLE beacons, there is no risk of identifying specific occupants. They can serve thus during the current time as well as in the post pandemic new normal. 
BLE of many form factors and sizes are available on the market.
Some can be installed on a key ring or other have the same size of a credit or access card. 

We propose the coupling of UVC luminaires and a multilayered occupancy detection system as a viable alternative for safe, automated decontamination of workspaces.
Our system has the capabilities of possibly automatically disinfecting the space multiple times over a work-day. 
Our system can be easily integrated in new and existing building and presents itself as a valuable addition to enhance the resilience of buildings and their healthy operation. 
We have made our entire work open source to encourage further development and free access to everyone, thus striving towards more equitable distribution of our findings and the knowledge created.

\section*{Acknowledgement}
This research was funded by the Republic of Singapore's National Research Foundation through a grant to the Berkeley Education Alliance for Research in Singapore (BEARS) under their Central Gap Fund, ``To thrive in the new Covid-19 normal'' scheme. The research was carried out within the state of the art SinBerBEST (Singapore-Berkeley Building Efficiency and Sustainability for Tropics) Testbed (\url{http://sinberbest.berkeley.edu/content-page/testbed-facilities}).

\bibliographystyle{IEEEtran}
\bibliography{references}

\end{document}